\documentclass[sigconf]{acmart}
\copyrightyear{2019}
\acmYear{2019}
\acmConference[MM '19]{Proceedings of the 27th ACM International Conference on Multimedia}{October 21--25, 2019}{Nice, France}
\acmBooktitle{Proceedings of the 27th ACM International Conference on Multimedia (MM '19), October 21--25, 2019, Nice, France}
\acmPrice{15.00}
\acmDOI{10.1145/3343031.3350933}
\acmISBN{978-1-4503-6889-6/19/10}

\def\BibTeX{{\rm B\kern-.05em{\sc i\kern-.025em b}\kern-.08emT\kern-.1667em\lower.7ex\hbox{E}\kern-.125emX}}

\usepackage{booktabs}
\usepackage{multirow}
\usepackage{array}
\usepackage{bigstrut}
\usepackage{balance}
\newcolumntype{L}[1]{>{\raggedright\arraybackslash}m{#1}}
\newcolumntype{C}[1]{>{\centering\arraybackslash}p{#1}}

\begin{document}
\fancyhead{}
%
\title{Eye in the Sky: Drone-Based Object Tracking and 3D Localization}

%
\author{Haotian Zhang}
\authornotemark[1]
\email{haotiz@uw.edu}
\affiliation{%
  \institution{University of Washington}
  \city{Seattle}
  \state{Washington}
  \postcode{98105}
}

\author{Gaoang Wang}
\email{gaoang@uw.edu}
\affiliation{%
  \institution{University of Washington}
  \city{Seattle}
  \state{Washington}
  \postcode{98105}
}

\author{Zhichao Lei}
\email{zl68@uw.edu}
\affiliation{%
  \institution{University of Washington}
  \city{Seattle}
  \state{Washington}
  \postcode{98105}
}

\author{Jenq-Neng Hwang}
\email{hwang@uw.edu}
\affiliation{%
  \institution{University of Washington}
  \city{Seattle}
  \state{Washington}
  \postcode{98105}
}

%
\renewcommand{\shortauthors}{H. Zhang, et al.}

%
\begin{abstract}
    Drones, or general UAVs, equipped with a single camera have been widely deployed to a broad range of applications, such as aerial photography, fast goods delivery and most importantly, surveillance. Despite the great progress achieved in computer vision algorithms, these algorithms are not usually optimized for dealing with images or video sequences acquired by drones, due to various challenges such as occlusion, fast camera motion and pose variation. In this paper, a drone-based multi-object tracking and 3D localization scheme is proposed based on the deep learning based  object detection. We first combine a multi-object tracking method called TrackletNet Tracker (TNT) which utilizes temporal and appearance information to track detected objects located on the ground for UAV applications.  Then, we are also able to localize the tracked ground objects based on the group plane estimated from the Multi-View Stereo technique. The system deployed on the drone can not only detect and track the objects in a scene, but can also localize their 3D coordinates in meters with respect to the drone camera. The experiments have proved our tracker can reliably handle most of the detected objects captured by drones and achieve favorable 3D localization performance when compared with the state-of-the-art methods.
\end{abstract}

%
%


%

%

%
\maketitle

\begin{figure}[t]
\begin{center}
\includegraphics[width=\linewidth]{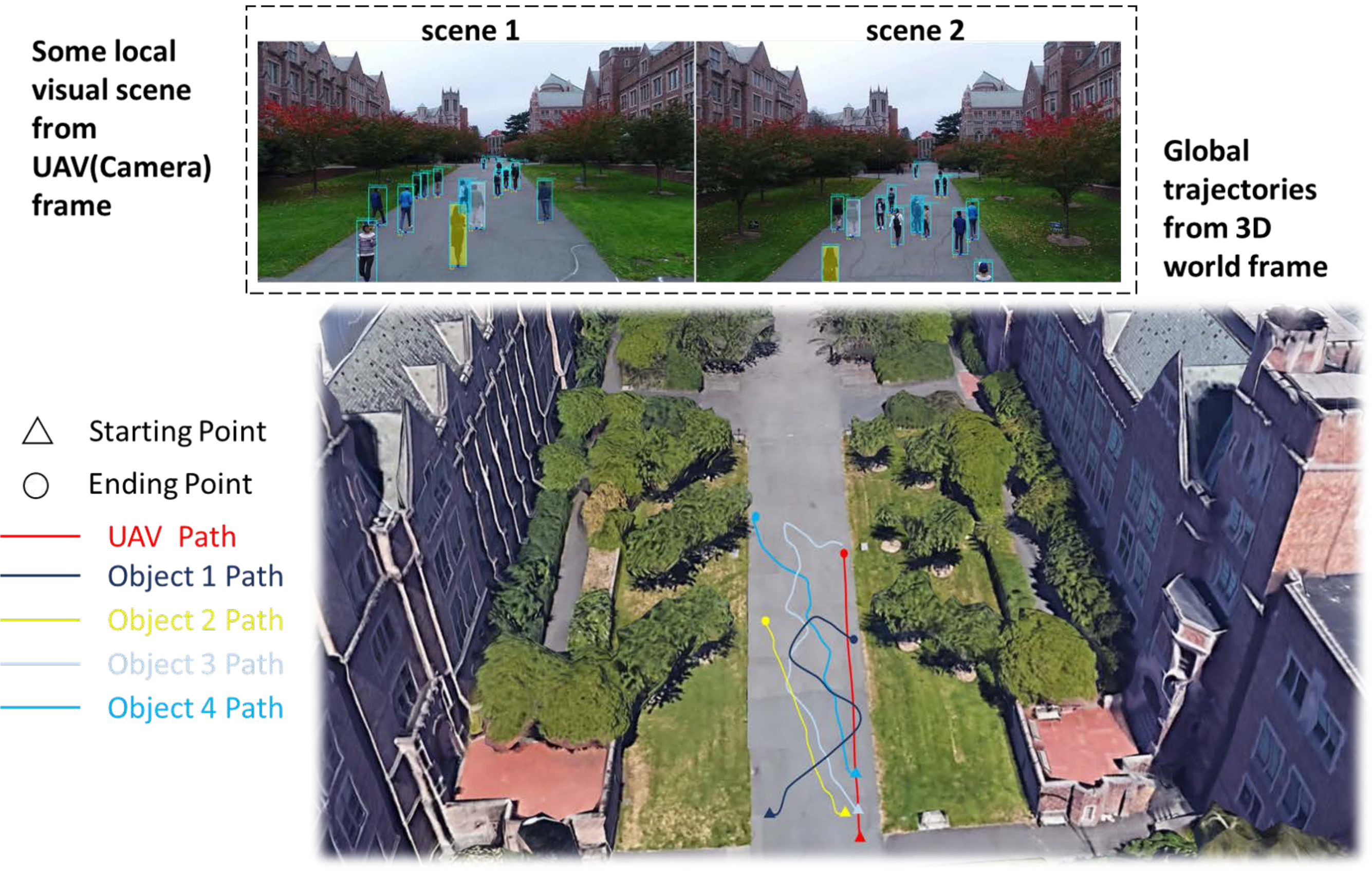}
\end{center}
   \caption{Demonstration of the UAV tracking and localization system. Some local visual scene in UAV (camera) frame and global trajectories in 3D world frame is shown. Each path of the object in the recorded frames begins from the start point towards the endpoint and different colors represent the different object.
   }
\label{fig:overall-flow}
\end{figure}

\section{Introduction}

\begin{figure*}
\begin{center}
\includegraphics[width=0.75\linewidth]{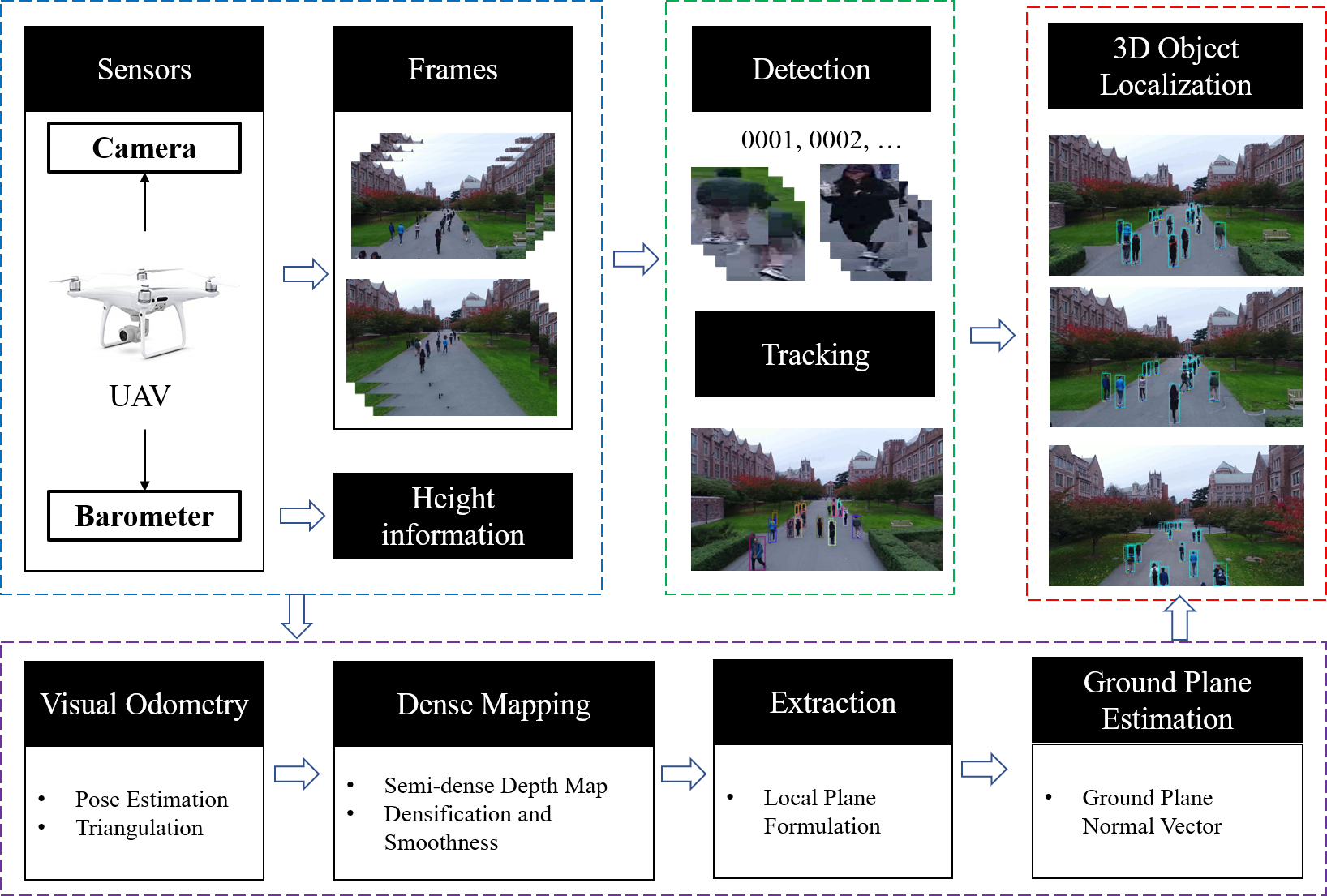}
\end{center}
   \caption{The flow chart of our proposed system, which integrates object detection, multi-object tracking and 3D localization.}
\label{fig:implent_flowchart}
\end{figure*}

Machine vision systems, such as monocular video cameras, and algorithms represent essential tools for several applications involving the use of unmanned aerial vehicles (UAVs). These techniques are frequently used to extract information about the surrounding scenes for several civilian/military applications like human surveillance, expedition guidance, and 3D mapping. Indeed, UAVs have the potential to dramatically increase the availability and usefulness of an aircraft as information-gathering platforms.

In addition to video cameras, multi-UAV missions have exploited various relative sensing systems for information gathering, such as Radio-Frequency (RF)-based ranging \cite{pupillo2015medicina}, LIDAR-based ranging\cite{lin2011mini}, etc. The main advantages relevant to vision systems are: (1) no additional sensors are needed; (2) visual cameras are extremely small, light and inexpensive with respect to the other sensors; (3) cameras can provide accurate line-of-sight information, which is often required in specific applications.

The novelties of the proposed system and the advantages are detailed below:

\begin{itemize}
    \item \textbf{Accurate object detection}: The proposed system detects objects of interest based on the modified RetinaNet \cite{lin2017focal}, which provides a better prior for tracking by detection \cite{kalal2012tracking} method compared with other state-of-the-art detectors.
    \item \textbf{Multi-object tracking}: A robust TrackletNet Tracker (TNT) for multiple object tracking (MOT), which takes into account both discriminative CNN appearance features and rich  temporal information, is incorporated to reduce the impact from unreliable or missing detections and generate smooth and accurate trajectories of moving objects.
    \item \textbf{Visual odometry and ground plane estimation}: We use the effective semi-direct visual odometry (SVO) \cite{forster2014svo} to get the camera pose between views. The ground plane is then estimated from dense mapping based on the multi-view stereo (MVS) \cite{seitz2006comparison} method. It minimizes photometric errors across frames and uses a regularization term to smoothen depth map in low-textured region.
    \item \textbf{3D object localization}: Based on the self-calibrated drone camera parameters, available camera height and estimated ground plane, the detected and tracked objects can be back-projected to 3D world coordinates from 2D image plane. The distance between objects and drones can thus be obtained.
\end{itemize}

The rest of the paper is organized as follows: Section \ref{sec:related} provides an overview of related works, with a focus on drone-based vision techniques. Furthermore, the originality and the advantages of each proposed module are motivated and addressed. Sections \ref{sec:proposed_system} presents the practical contexts in which every part of the proposed tracking and 3D ground object localization system are developed. Section \ref{sec:experiments} provides detailed implementation details and extensive experiments results to show the accuracy and robustness of our systems, followed by the conclusions in Section \ref{sec:conclusion}.

\section{Related Work}
\label{sec:related}

The key enabling technologies required in the cognitive task of our proposed drone-based system mainly include object of interest detection, multiple object tracking, detected object 3D localization and overall system integration. In this section, we will present a review of related works on each of modules and open issues ahead.
\paragraph{\textbf{Object of Interest Detection.}}
Most of surveillance drones fly with low-altitude, so that the ground objects to be detected are within the range of views. Existing vision approaches for object detection are classified into two categories: (1) direct and feature-based methods, and (2) deep learning methods. The latter methods usually achieve higher performance and now become the state-of-the-art techniques. Among which, Faster R-CNN \cite{ren2015faster}, SSD \cite{liu2016ssd}, YOLO \cite{redmon2016you} and RetinaNet \cite{lin2017focal} are the most popular deep learning detectors used by researchers. However, due to the critical challenges such as fast camera motions, occlusion and relative motion between camera and targets that can cause a significant and high-dynamic variation of sudden appearance changes, the above mentioned deep learning detectors may not be optimal for such scenarios.

\paragraph{\textbf{Multiple Object Tracking.}}
Most of the recent multi-object tracking (MOT) methods are based on tracking-by-detection schemes\cite{sadeghian2017tracking}. Given detection results, we are able to associate detections across frames and locate objects in 2D even when unreliable detections and occlusions occur. Common tracking frameworks, such as the Graph Model proposed in \cite{milan2016multi}, try to solve the problem by minimizing the total energy loss. However, using graph models for representation requires the nodes (detections) to be conditionally independent, which is usually not the case. Some other frameworks, such as Tracking by Feature Fusion \cite{ristani2018features, zhang2017multi, tang2017multiple}, usually jointly fuse classical features (HOG, color histogram and LBP) as appearance features and locations/speed of 2D bounding boxes from detections as temporal features, nonetheless, it is still hard and quite heuristic to determine each weight for feature fusion. Other approaches, like End-to-End deep learning based Tracking \cite{feichtenhofer2017detect,kang2017t,kang2016object}, can sometimes be successful but require huge amount of labelled training data. It is usually not the case for drones since it is very laborious for human labelling of tiny objects in the drone videos.

\paragraph{\textbf{Ground Object Localization.}} There are two definitions for 3D localization in our paper: (1) Self-localized (self-calibrated) of the camera extrinsic parameters to get its own world positions; (2) 3D-localized of detected target objects and get the distance from camera to objects;

To achieve the first goal, simultaneous localization and mapping (SLAM) technology is introduced. ORB-SLAM \cite{mur2017orb} is a symbolic framework for computing camera trajectory in real-time by extracting and tracking feature points across video frames and reconstruct sparse point cloud using camera geometry. Foster et al. \cite{forster2014svo} propose a semi-direct monocular visual odometry (SVO), which uses pixel brightness to estimate pose, resulting in the ability to maintain pixel-level based precision in high-frame-rate video, and can generate denser map compared to the ORB-SLAM. To accomplish goal of the second definition, as we all know, it is impossible for an object from a single image to obtain the distance of the object to camera. Knoppe et al. \cite{honkavaara2013processing} propose a system for a drone carrying a stereo camera to get ground surface scanning data. Karol et al. \cite{mikadlicki2017real} use a drone-mounted LIDAR to do the ground plane estimation. Nonetheless, the use of additional cameras and advanced sensors could generate problems for a drone such as the increased payload, etc. Thus, the obvious solution is to carry a single camera.

Traditional computer vision techniques also indicate that a ground object can be accurately 3D-localized if the camera pose, its height and ground plane patch beneath the object is known.

\paragraph{\textbf{Overall System Integration.}} There are few works have been done for high-level drone-based surveillance systems. Surya et al. \cite{penmetsa2014autonomous} propose an autonomous drone surveillance system that can detect individuals engaged in violent activities. Singh et al. \cite{singh2018eye} use a feature pyramid network (FPN) as the object detector and a ScatterNet Hybrid Deep Learning (SHDL) Networks to estimate the pose of each detected human. However, both works are still very much in its early stage and all their techniques have been demonstrated only based on 2D coordinates. In real world applications, a much better way for a drone to achieve surveillance aim is to infer where are the ground targets (distances) in 3D world space (meters) and how they will move in the future, so that some actions can be predicted according to targets' locations, movements, and speed, etc.

\section{Proposed Tracking and Localization System}
\label{sec:proposed_system}

\subsection{TrackletNet based MOT Tracker}
\label{sec:mot}

As shown in Figure~\ref{fig:implent_flowchart}, our proposed drone-based multiple object tracking (MOT) and 3D localization system only require one single monocular video camera, which can systematically and dynamically calibrate its own extrinsic parameters in order to achieve the self-localization. The ground plane in the view is then estimated by a multi-view stereo\cite{seitz2006comparison} method to infer 3D coordinate transformation of the image pixels. Based on the calibrated camera parameters and estimated ground plane, the detected and tracked objects of interests (pedestrians, cars) on the ground can be 3D localized in either the image or the world coordinates.

\begin{figure}[t]
\begin{center}
\includegraphics[width=\linewidth]{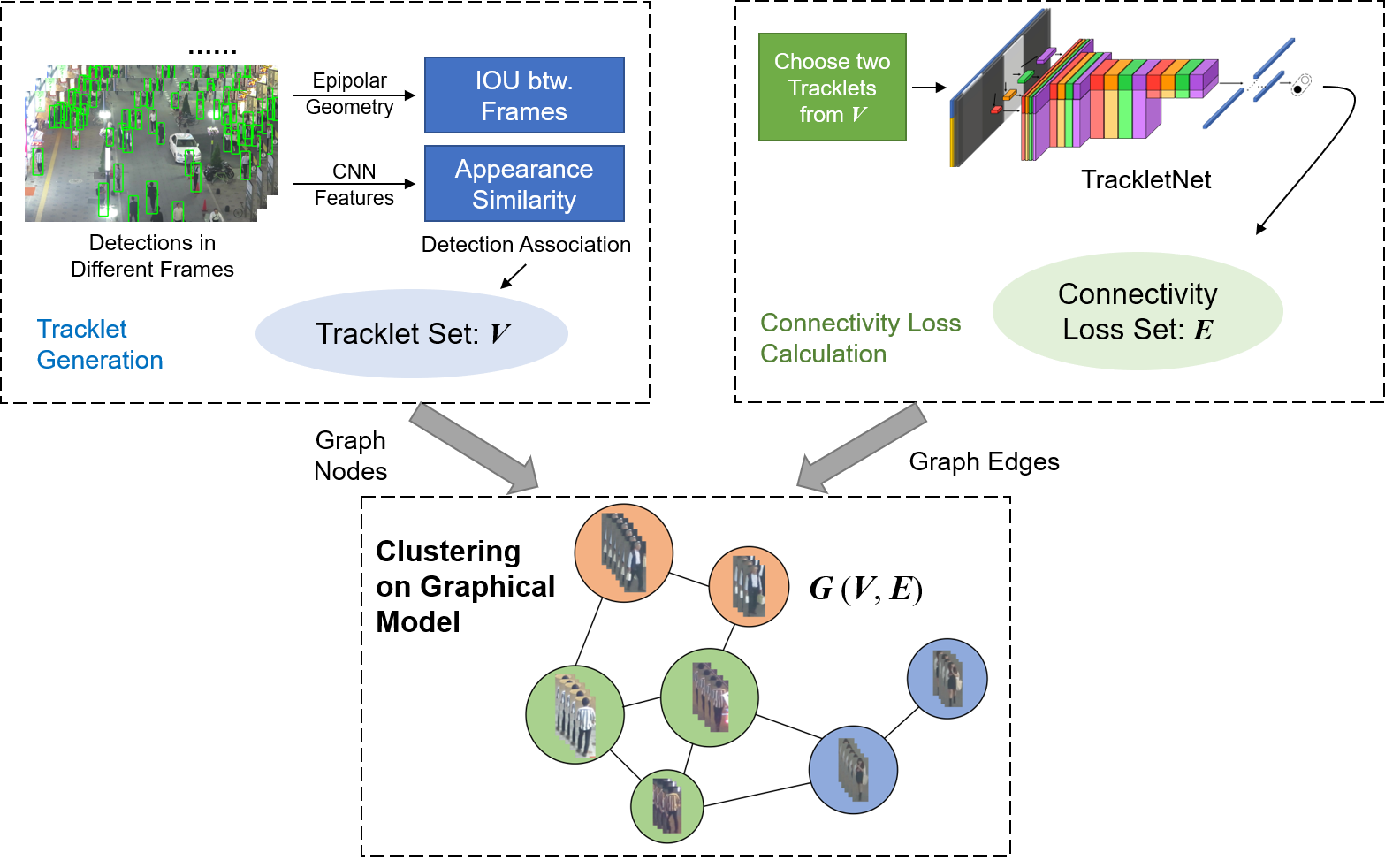}
\end{center}
   \caption{The TNT framework for multi-object tracking. Given the detections in different frames, detection association is computed to generate Tracklets for the Vertex Set $V$. After that, every two tracklets are put into the TrackletNet to measure the degree of connectivity, which form the similarity on the Edge Set $E$. A graph model $G$ can be derived from $V$ and $E$. Finally, the tracklets with the same ID are grouped into one cluster using the graph partition/clustering approach.
   }
\label{fig:tnt}
\end{figure}

We adopt TrackletNet Tracker (TNT) \cite{wang2018exploit} in our UAV applications. The tracking system is based on a tracklet graph-based model, as shown in Figure~\ref{fig:tnt}, which has three key components, 1) tracklet generation, 2) connectivity measure, and 3) graph-based clustering. Given the detection results in each frame, each tracklets to be treated as a node in the graph is generated based on the intersection-over-union (IOU) compensated by the epipolar geometry constraint due to camera motion and the appearance similarity between two adjacent frames. Between every two tracklets, the connectivity is measured as the edge weight in the graph model, where the connectivity represents the likelihood of the two tracklets being from the same object. To calculate the connectivity, a multi-scale TrackletNet is built as a classifier, which can combine both temporal and spatial features in the likelihood estimation. Clustering \cite{tang2018single} is then conducted to minimize the total cost on the graph. After clustering, the tracklets from the same ID can be merged into one group.

The reason we use TNT as our tracking method is due to its robustness dealing with erroneous detections caused by occlusions and missing detections. More specifically, 1) The TrackletNet focuses on the continuity of the embedded features along the time. In other words, the convolution kernels only capture the dependency along time. 2) The network integrates object Re-ID, temporal and spatial dependency as one unified framework. Based on the tracking results from TNT, we know the continuous trajectory of each object ID across frames. This information will be used in the object 3D localization to be discussed in the subsequent subsection.

\subsection{Semi-Direct Visual Odometry}
\label{sec:svo}

\begin{figure}[t]
\begin{center}
\includegraphics[width=0.7\linewidth]{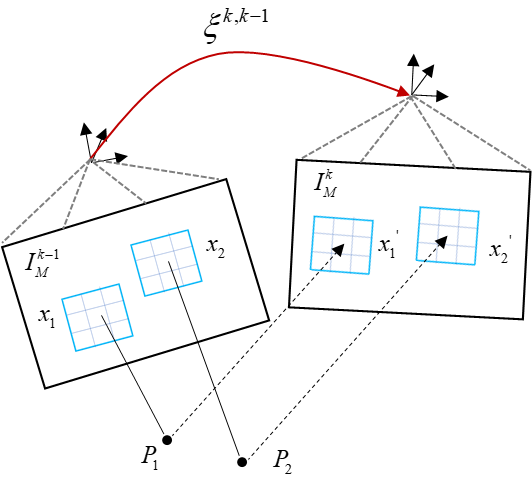}
\end{center}
   \caption{Changing the relative pose ${\xi ^{k,k - 1}}$ between the current and the previous frame implicitly moves the position of the reprojected points in the new image. Sparse image alignment seeks to find ${\xi ^{k,k - 1}}$  that minimizes the photometric difference between image blocks corresponding to the same 3D point (blue blocks) corresponding to the same 3D point ($P_1, P_2$).
   }
\label{fig:svo}
\end{figure}

To self-calibrate the drone camera, i.e., to estimate the extrinsic camera parameters frame-by-frame, we use a monocular semi-direct visual odometry (SVO) algorithm \cite{forster2014svo, schops2014semi}, which directly operates on the raw intensity image instead of using extracted features at any stage of the algorithm. As shown in Figure~\ref{fig:svo}, we represent the image as function $I: \Omega  \to \mathbb{R}$. Similarly, we represent the inverse depth map and inverse depth variance as functions $D:{\Omega _D} \to {\mathbb{R}^ + }$ and $V:{\Omega _D} \to {\mathbb{R}^ + }$, where ${\Omega _D}$ contains all the pixels which should have a valid depth hypothesis. Note that $D$ and $V$ separately denote the mean and variance of the \emph{inverse depth}, which is assumed as a Gaussian-distributed depth. The depth values of extracted SIFT \cite{lowe2004distinctive} feature points are initialized with random depth values and large variance for the first frame. Assume the camera moves slowly and in parallel to the image plane, the SVO will quickly converge to a valid map. The pose of a new frame is then estimated using direct image alignment, more specifically, given the current map $\{ {I_M},{D_M},{V_M}\} $, the relative pose $\xi  \in SE(3)$  of a new frame $I$ is obtained by directly minimizing the photometric error.
\begin{equation}
    E(\xi ): = \sum\limits_{x \in {\Omega _{{D_M}}}} {{{\left\| {{I_M}(x) - I(\omega (x,{D_M}(x),\xi ))} \right\|}_\delta },}
    \label{eq:svo}
\end{equation}
where $\omega :{\Omega _{{D_M}}} \times \mathbb{R} \times SE(3) \to \Omega $  projects a point from reference image plane to the new frame, and ${\left\|  \cdot  \right\|_\delta }$ is the \emph{Huber} norm to account for outliers.

In order to make the approach more robust, we propose to aggregate the photometric cost in a small pixel block centered at the feature pixel and approximate the neighboring pixels as those estimated for the SIFT feature points. The minimization is computed using standard nonlinear least squares algorithms, such as Levenberg-Marquardt (LM).

\subsection{Depth Map from Multi-View Stereo}
\label{sec:depth_multi_view}


Based on the sparse depth map values estimated from SVO, we further formulate the dense depth calculation as a Gaussian estimation problem \cite{pizzoli2014remode} so as to estimate the depth map values surrounding the initialized SIFT points based on multiple frames of a monocular video. As discussed in Section \ref{sec:svo}, the relative pose between subsequent frames and the depth at semi-direct feature locations are estimated from SVO. Each observation gives a depth measurement by triangulating from the reference view and the last acquired view. The depth of a pixel block can be continuously updated on the basis of the current observation. Finally, densification and smoothness on the resulting depth map based on multiple observations is achieved.

\begin{figure}[t]
\begin{center}
\includegraphics[width=0.8\linewidth]{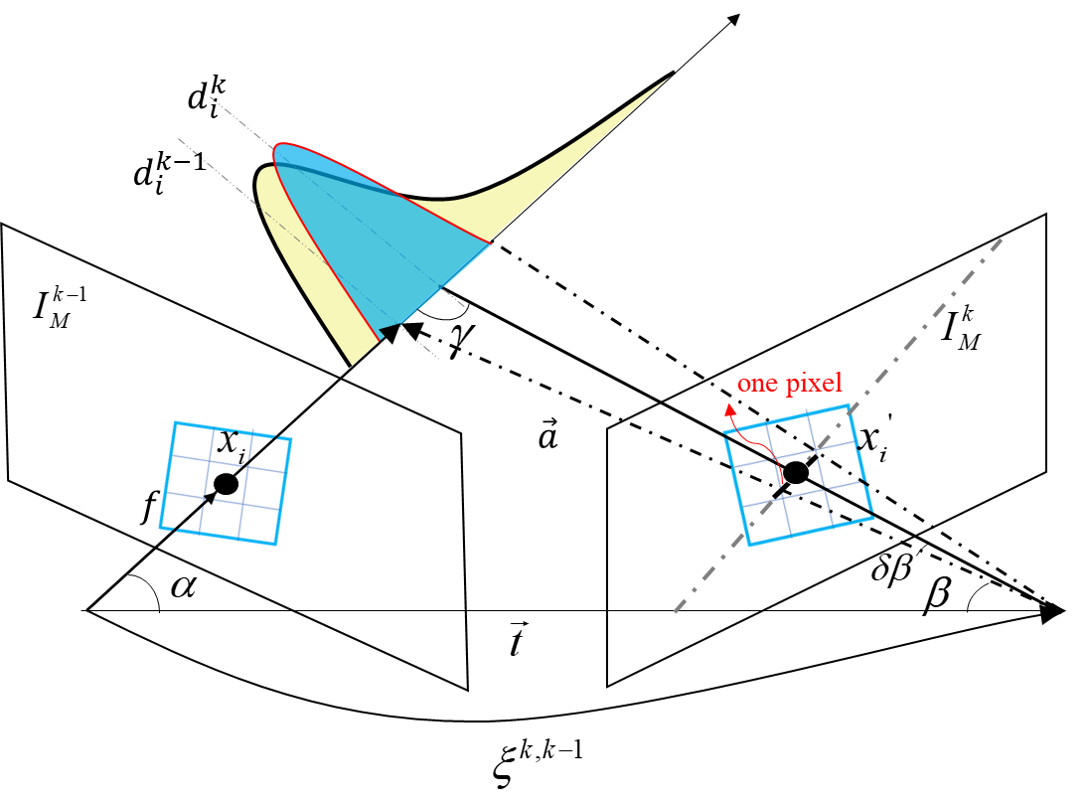}
\end{center}
   \caption{Probabilistic depth estimate $d_i^k$ for feature $i$ in the reference frame $I_M^{k - 1}$. The point at the true depth projects to similar image regions in both images (blue squares). The point of highest correlation lies always on the epipolar line in the new image.
   }
\label{fig:depth_filter}
\end{figure}

More specifically, for a set of previous keyframes as well as every subsequent frame with known relative camera pose, a block matching epipolar search is performed to search for the highest correlation. Several metrics to describe the similarities can be introduced to form the block matching problem, such as Sum of Absolute Distance (SAD) \cite{lempitsky2010learning}, Sum of Squared Distance (SSD) \cite{weinberger2006distance}, and Normalized Cross Correlation (NCC) \cite{lowe2004distinctive}, among which NCC has been commonly used as a metric to evaluate the degree of similarity between two compared pixel blocks. The main advantage of the NCC is that it is less sensitive to linear changes in the amplitude of illumination in two compared pixel blocks. In our case, the block matching between the block centered at ${x_i}$ in frame $I_M^{k - 1}$ and that of ${x_i}^{'}$ in frame $I_M^k$ can be given as,

\begin{equation}
    S({x_i},{x_i}^{'}) = \frac{{\sum\nolimits_{m,n} {{x_i}(m,n){x_i}^{'}(m,n)} }}{{\sqrt {\sum\nolimits_{m,n} {{x_i}{{(m,n)}^2}{x_i}^{'}{{(m,n)}^2}} } }}
    \label{eq:block_match}
\end{equation}

where $(m,n)$ indicates each pixel inside the corresponding block. If the resulting value is close to 1, which means two pixel blocks between two consecutive frames are very likely to be the same. The problem might occur if the epipolar search is long or the block becomes non-textured, we are very likely to encounter a non-convex distribution for correlation score, resulting in a very unreliable and non-smooth depth map. However, we always know that this is a one-to-one problem, therefore the depth filter is thus introduced for further processing.

We model the depth filter based on a Gaussian distribution, which is the depth $d$ (D) (normally distributed around the true depth). Hence, the probability of depth measurement $d_i^k$ for each block $i$ at frame $k$ is modeled as:

\begin{equation}
    p(d_i^k) \sim N(d_i^k|{\mu _i},\sigma _i^2)
    \label{eq:depth_filter}
\end{equation}

where ${\mu _i}$ represents the mean and ${\sigma _i}^2$ represents variance of the performance of Gaussian distribution of depth measurement, whose parameters could be estimated in a maximum likelihood framework using Expectation Maximization. Since each observation gives a depth measurement by triangulating from reference view and the last acquired view, given the consecutively multiple independent observations \{  ${d^k}$ , for $k = 1,2,...,N$ \}, the depth estimation can be continuously refined by Bayesian propagation, i.e,

\begin{equation}
    p(\mu ,{\sigma ^2}|{d^1},...,{d^N}) \propto p(\mu ,{\sigma ^2})\prod\limits_k {p({d^k}|\mu ,{\sigma ^2})}
    \label{eq:propagation}
\end{equation}

where $p(\mu ,{\sigma ^2})$ is our prior on depth. The ${\mu ^k}$ and ${\sigma ^k}$ can be iteratively obtained from relative positions of the camera at frame $k - 1$ and $k$. According to Figure~\ref{fig:depth_filter}, let  $\vec t$ be the translation component of relative pose $\xi $ and $f$ be the camera focal length, $\left\| {{{\vec d}^{k - 1}}} \right\|$ and $\left\| {\vec a} \right\|$ ($ \gg f$ ) are the depth regarding to image frame $I_M^{k - 1}$ and $I_M^k$ , which are obtained from triangulation, then:
\begin{equation}
    \alpha  = arccos\left( {\frac{{{{\vec d}^{k - 1}} \cdot \vec t}}{{\left\| {{{\vec d}^{k - 1}}} \right\|\left\| {\vec t} \right\|}}} \right)
\end{equation}
\begin{equation}
    \beta  = arccos\left( {\frac{{\vec a \cdot ( - \vec t)}}{{\left\| {\vec a} \right\|\left\| { - \vec t} \right\|}}} \right)
\end{equation}
Let $\delta \beta$ be the angle spinning for one pixel:
\begin{equation}
    \delta \beta  = \arctan \frac{1}{f}
\end{equation}
\begin{equation}
    \gamma  = \pi  - \alpha  - (\beta  + \delta \beta )
\end{equation}
Applying the law of sines, we can recover the norm of the updated ${\vec d^k}$  :
\begin{equation}
   \left\| {{{\vec d}^k}} \right\| = \left\| t \right\|\frac{{\sin (\beta  + \delta \beta )}}{{\sin \gamma }}
\end{equation}

Hence, the ${\mu ^k}$ and ${\sigma ^k}$ can be represented as
\begin{equation}
\begin{aligned}
    {\mu ^k} = \frac{1}{2}(\left\| {{{\vec d}^{k - 1}}} \right\| + \left\| {{{\vec d}^k}} \right\|) \\
    {\sigma ^k} = \left\| {{{\vec d}^k}} \right\| - \left\| {{{\vec d}^{k - 1}}} \right\| \\
\end{aligned}
\end{equation}
By using Eq.~\ref{eq:propagation}, the estimates of ${\mu ^k}$ and ${\sigma ^k}$ will eventually converge to the correct value and the depth is updated on the basis of the current observation.

For densification, we extend PatchMatch Stereo \cite{bleyer2011patchmatch} to Multiview form. We keep the camera poses from SVO and epipolar search for best depth value for each local block. Search and updating for the best value for each block is time-consuming, however, PatchMatch uses a belief propagation to accelerate the updating process. For each block, we look for the depth value with least photometric error and propagate it to the other neighboring pixels using bilinear interpolation.

\subsection{3D Object Localization via Ground Plane Estimation}
\label{sec:3d_object_localization}

\begin{figure}[t]
\begin{center}
\includegraphics[width=0.9\linewidth]{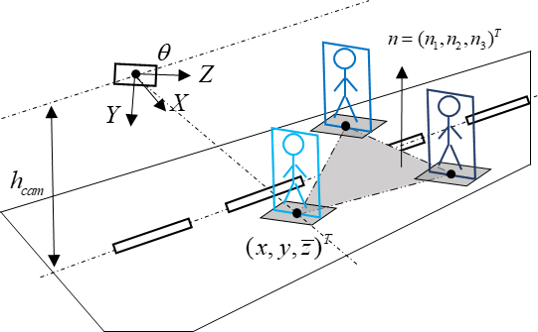}
\end{center}
   \caption{Coordinate system definitions for 3D object localization. The ground plane is defined as a $\{ {n^T},{h_{cam}}\}$. $\bar z$ is the average depth of the surrounding area.
   }
\label{fig:gpe}
\end{figure}

As shown in Figure~\ref{fig:gpe}, the camera height above the ground ${h_{cam}}$ is defined as the distance from the principle center to the ground plane. For a common geometry representation of the ground plane, the ground plane is defined as ground height ${h_{cam}}$ and the unit normal vector $n = {(n1,n2,n3)^T}$ . There exists a pitch angle $\theta $ between the drone and ground plane. For any 3D point ${(x,y,z)^T}$ on the ground plane, we have ${h_{cam}} = y\cos \theta  - zsin\theta $.

Assume we obtain the depth map from \ref{eq:propagation} and there are multiple objects on the ground, we use the multiple average depth values {$\bar z$} surrounding the bottom center points of each bounding boxes of multiple detected objects to form a local plane. Once such a plane is obtained, we can get the unit normal vector $n = {(n1,n2,n3)^T}$ by using Cramer \textquotesingle s rule \cite{kyrchei2010cramer}:

\begin{equation}
    \begin{aligned}
    {n_1} = \sum {y\bar z}  \times \sum {xy - } \sum {x\bar z}  \times \sum {yy}  \\
    {n_2} = \sum {xy}  \times \sum {x\bar z - } \sum {xx}  \times \sum {y\bar z}  \\
    {n_3} = \sum {xx}  \times \sum {yy - } \sum {xy}  \times \sum {xy}
\end{aligned}
\end{equation}

\paragraph{\textbf{3D Object Localization.}}
Accurate estimation of both ground height and orientation is crucial for 3D object localization \cite{song2015joint}. Let $K$ be the camera intrinsic calibration matrix. The bottom center of a 2D bounding box, $b = {(x,y,1)^T}$ in homogeneous coordinates, can now be back-projected to 3D through the ground plane $\{ {n^T},{h_{cam}}\}$.
\begin{equation}
    c = \pi _G^{ - 1}(b) = \frac{{{h_{cam}}{K^{ - 1}}b}}{{{n^T}{K^{ - 1}}b}}
\end{equation}

\section{Experiments}
\label{sec:experiments}

\subsection{Datasets}
\label{sec:datasets}
Two datasets are used to evaluate our performance for each stage.
\paragraph{\textbf{VisDrone-2018.}}
VisDrone benchmark dataset \cite{zhu2018vision} was proposed at ECCV 2018 workshop. The benchmark datasets consist of 263 video clips with 179,264 frames, captured by various drone-mounted cameras. Objects of interests frequently appear in the image are pedestrians, cars, buses, etc. Tasks involved in this dataset, such as object detection and multi-object tracking, are extremely challenging due to issues such as occlusion, large scale, and pose variation and fast motions. The dataset is used to evaluate our performance for detection and tracking modules.


\paragraph{\textbf{Our own-recorded dataset.}}
We chose the commercial UAV DJI Phantom 4 as a platform for the data acquisition. The video frames were captured by the equipped monocular camera, which guarantees high-quality video/image acquisition during speed movement with its wide-angle fixed focal length, and a shooting screen without distortion. The barometer module on the drone is used to measure the flight attitude for the monocular visual odometry scale correction. Our own datasets also cover different environments, including campus, grass land, basketball field, etc. The target objects positions are recorded using hand-hold GPS device. We then human-labeled the positions by refining them into multiple grids. Finally, all trajectories are smoothened and can be regarded as the ground truth.

\subsection{Implementation Details}
\label{sec:implementation}

\paragraph{\textbf{Object Detection.}}
Our trained detector was based on the RetinaNet50 Detector \cite{lin2017focal, zhu2018visdrone}. We changed the anchor size to detect smaller objects. For the same reason, we added a \texttt{CONV} layer in FPN's \texttt{P3} and \texttt{P4}, where the higher-level features are added to the lower-level features. We also used the multi-scale training techniques and the Soft-NMS \cite{bodla2017soft} algorithm in post processing. The detector was pretrained on MOT16 \cite{milan2016mot16} and fine-tuned on VisDrone2018-DET datasets. We split the training datasets from VisDrone-2018-DET into 6,000 frames for training and 1,048 frames for testing. We evaluated our detection performance for only pedestrians, cars and buses after 20,000 epochs. The mAP for each class reached 86.2\%, 97.8\%, 95.5\% respectively.

\begin{table*}
\begin{center}
\begin{tabular}{|L{2.9cm}|C{1.3cm}|C{1.2cm}|C{1.2cm}|C{1.2cm}|C{1.2cm}|C{1.2cm}|C{1.2cm}|}
\hline
Tracker & MOTA $\uparrow$ & IDF1 $\uparrow$ & MT $\uparrow$ & ML $\downarrow$ & FP $\downarrow$ & FN $\downarrow$ & IDsw. $\downarrow$ \\
\hline\hline
V\_IOU \cite{bochinski2018extending} & 40.2 & 56.1 & \noindent\color{cyan}{297} & 514 & 11,838 & 74,027 & \textbf{265} \\
TrackCG \cite{tian2015fast} & \noindent\color{cyan}{42.6} & \noindent\color{cyan}{58.0} & 323 & \noindent\color{cyan}{395} & 14,722 & \noindent\color{cyan}{68,060} & 779 \\
GOG\_EOC \cite{pirsiavash2011globally} & 36.9 & 46.5 & 205 & 589 & \noindent\color{cyan}{5,445} & 86,399 & 754 \\
SCTrack \cite{al2017robust} & 35.8 & 45.1 & 211 & 550 & 7,298 & 85,623 & 798 \\
Ctrack \cite{tian2017joint} & 30.8 & 51.9 & \textbf{369} & \textbf{375} & 36,930 & \textbf{62,819} & 1,376 \\
FRMOT \cite{ren2015faster} & 33.1 & 50.8 & 254 & 463 & 21,736 & 74,953 & 1,043 \\
GOG \cite{pirsiavash2011globally} & 38.4 & 45.1 & 244 & 496 & 10,179 & 78,724 & 1,114 \\
CMOT \cite{bae2014robust} & 31.5 & 51.3 & 282 & 435 & 26,851 & 72,382 & 789 \\
\hline
\textbf{Ours} & \textbf{48.6} & \textbf{58.1} & 281 & 478 & \textbf{5,349} & 76,402 & \noindent\color{cyan}{468} \\
\hline
\end{tabular}
\end{center}
\caption{Tracking performance on the VisDrone2018-MOT test set compared to state-of-the-art. Best in bold, second best in blue.
}
\label{tab:vis}
\end{table*}

\paragraph{\textbf{Multi-Object Tracking.}}
Similar to the training of the detector, we also pre-trained the multi-scale TrackletNet on MOT16 datasets, and then fine-tuned the model on VisDrone2018-MOT datasets. The VisDrone2018-MOT contains 56 video sequences for training (24,201 frames in total), and 33 sequences for testing. To generate better tracklets, the IOU\_threshold is set to 0.3 due to the drone's fast camera motions. The time window is set to 64 and batch size is set to 32. The Adam optimizer with an initial learning rate of 1e-3 and is decreased by 10 times for every 2,000 iterations.

\paragraph{\textbf{Intrinsic Camera.}}
The camera matrix $K$ is assumed to be known for every testing sequence. As we will show in the experiments, an approximation \cite{dragon2014ground} of focal length $f$.
\begin{equation}
    K = \left[ {\begin{array}{*{20}{c}}
f&0&{w/2}\\
0&f&{h/2}\\
0&0&1
\end{array}} \right] \text{ with} f = \frac{w}{2}\arctan (\frac{{{\rm{9}}0}}{{180}}\frac{\pi }{2})
\end{equation}
under an image size of $w \times h$, assuming a horizontal field of view of 90 degrees, is sufficiently accurate for drone-equipped cameras.

\paragraph{\textbf{Ground Plane Estimation.}}
As mentioned above, the camera pose estimation is based on semi-direct VO. The implementation is measured by an average drift in pose of 0.0045 meters per second for an average depth of 1 meter. We also estimated depth using sliding window approach by setting the window interval $N = 30$ frames. The area of ground patch beneath the object is chose to be $(a,\frac{1}{3}b)$, where $a$ is the width of the bounding box and $b$ is the height.

\subsection{Experimental Performance}
\label{sec:performance}

\paragraph{\textbf{Multi-Object Tracking on VisDrone2018-MOT datasets.}}

\begin{figure}[t]
\begin{center}
\includegraphics[width=\linewidth]{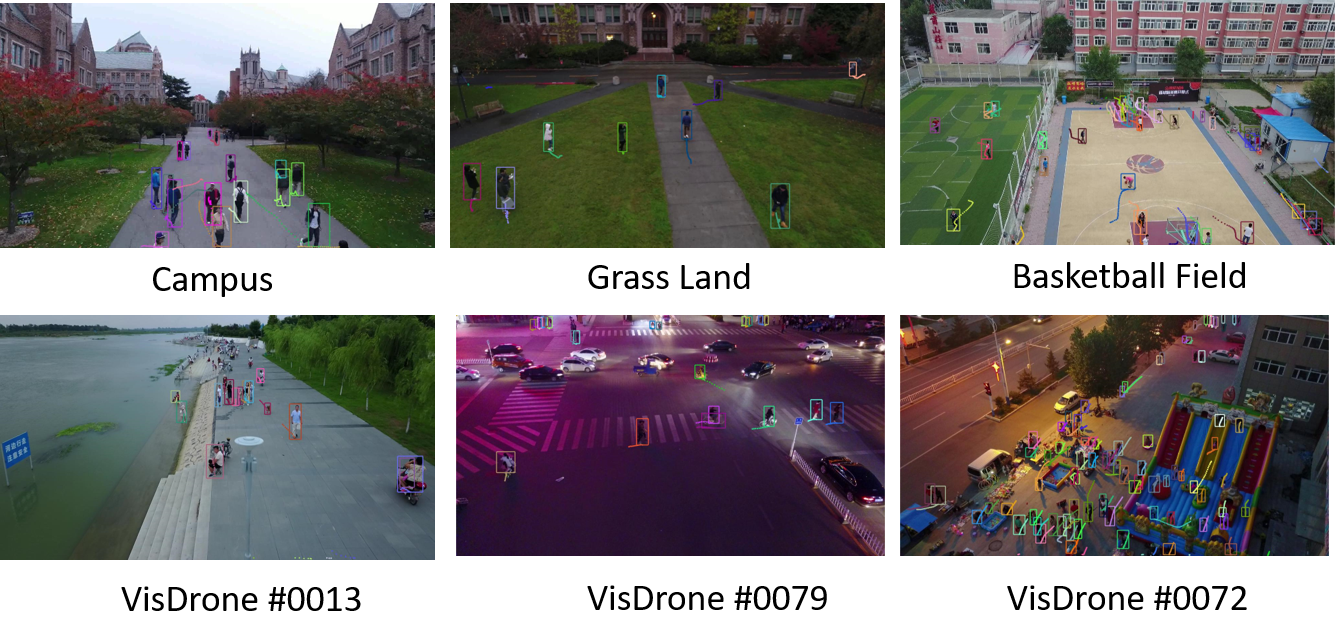}
\end{center}
   \caption{Tracking results on the test sequences in our recorded campus datasets and the VisDrone-MOT benchmark.}
\label{fig:track_result}
\end{figure}

We provide our qualitative results on VisDrone2018-MOT benchmark datasets by comparing with other state-of-the art methods, which are shown in Table~\ref{tab:vis}. Note that the benchmark datasets can evaluate performance on one of two different evaluation tasks, donated by without prior detection and with prior detection. As mentioned above, our method is based on tracking-by-detection, so the final performance is evaluated on provided Faster-RCNN detection results. Figure~\ref{fig:track_result} shows some examples of tracking results on both VisDrone dataset and our recorded datasets.

V\_IOU \cite{bochinski2018extending} is also a tracking by detection method, they assumed that the detections of an object in consecutive frames have an unmistakably high overlap IOU which is commonly the case when sufficiently high frame rates. However, their method is just a simple IOU tracker without incorporating the appearance information. TrackCG \cite{tian2017joint} proposed a novel approach by aggregating temporal events within target groups and integrating a graph-modeling based stitching procedure to handle the multi-object tracking problems. Yet, the graphical model is used for representation and requires the nodes (detections) be conditionally independent, which is usually not the case. Our method takes advantage of both appearance feature and temporal information into a unified framework based on an undirected graph model. By comparing the tracking performance, it can be seen that we achieved the first place on MOTA \cite{bernardin2008evaluating,milan2016mot16}, IDF1 \cite{ristani2016performance}, and FP (false positive).  Among these, IDF1 scores can effectively reflects how long of an object has been correctly tracked and MOT score computes the tracking accuracy. For other metrics like ID Switch, we are also among the top rankings.

\paragraph{\textbf{3D Localization Performance.}}

\begin{figure}[t]
\begin{center}
\includegraphics[width=0.8\linewidth]{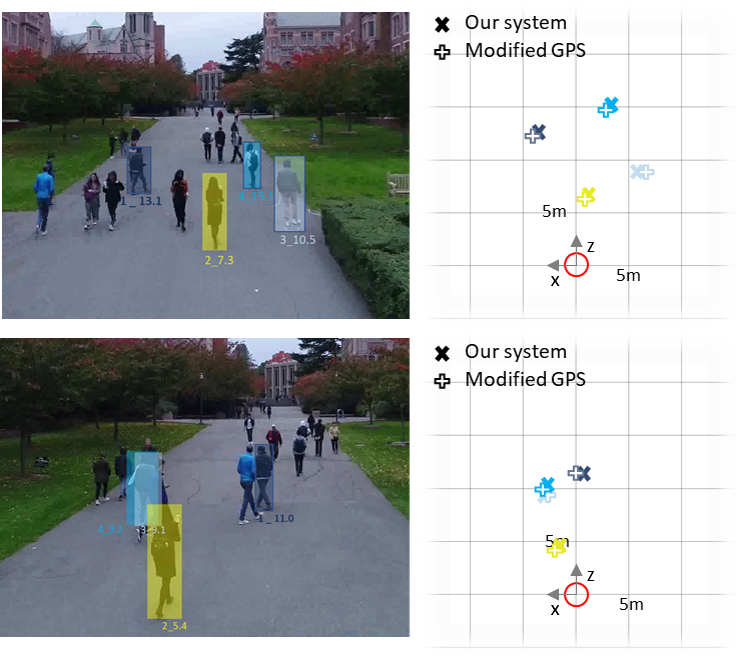}
\end{center}
   \caption{Output of our localization system. The left panel shows input 2D bounding boxes, its object id given by tracking and the estimated distance. The right panel shows the top view of the ground truth object localization from modified GPS results, compared to our 3D object localization given by our system.}
\label{fig:local_demo}
\end{figure}

\begin{figure}[t]
\begin{center}
\includegraphics[width=\linewidth]{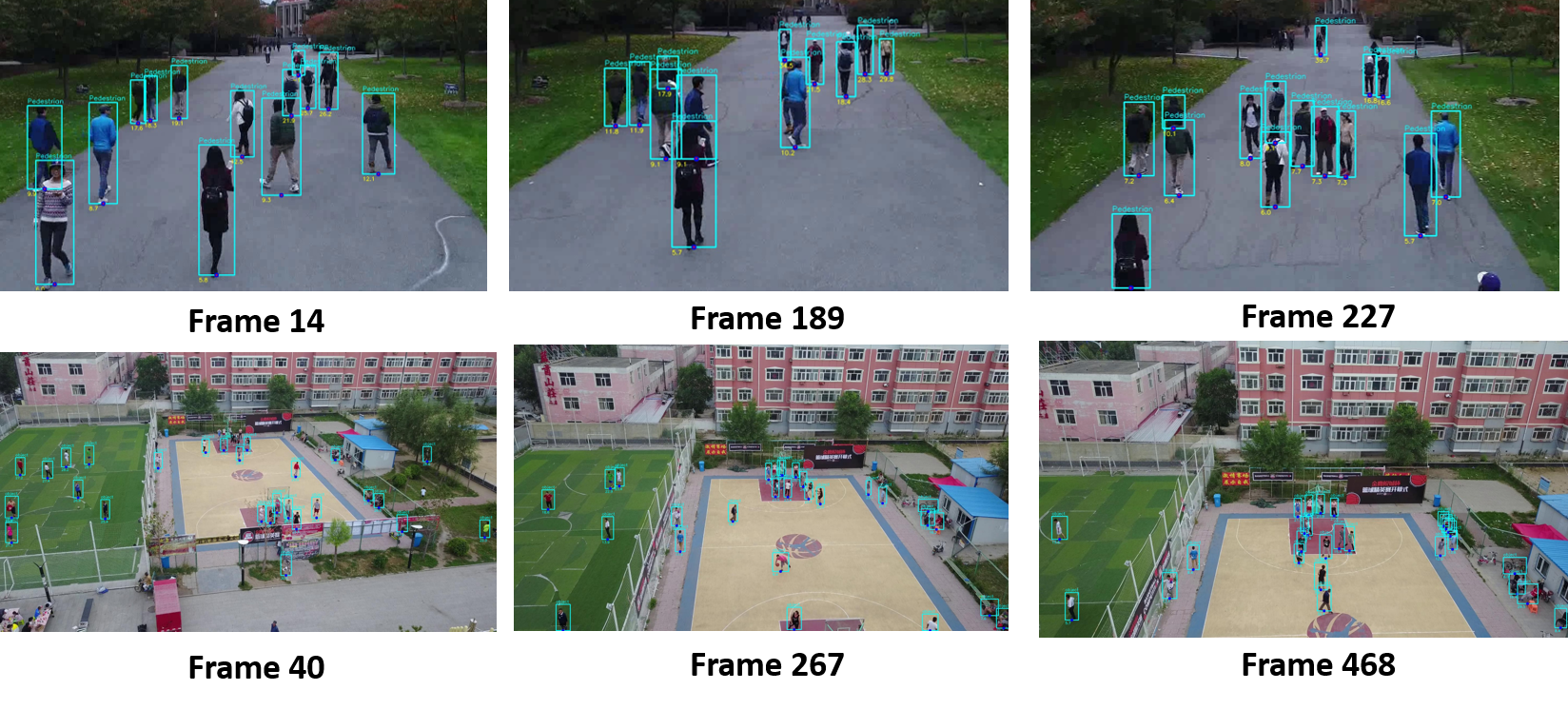}
\end{center}
   \caption{Sampled localization results. The distance between objects and the drone is displayed in yellow and white beneath the bounding boxes (Zoom in for better visualization).}
\label{fig:sample_loc}
\end{figure}

The output of our system is shown in Figure~\ref{fig:local_demo}. The 3D localization performance was evaluated under our captured sequences. As the drone flies to a higher altitude, or the object is farther away, the distance towards the object becomes less accurate. Some examples of 3D localization results are shown in Figure~\ref{fig:sample_loc}.



\begin{table*}
  \centering
  \caption{Mean localization error(standard deviation in parenthesis) in meters.}
    \begin{tabular}{|c|l|c|ccc|}
    \hline
    Approach & Scene & Overall (m) & $<=$10m & $<=$25m & $>$25m \bigstrut\\
    \hline
    \hline
    \multirow{3}[2]{*}{\texttt{Det+Flat\_Ground\_Asmp}} & Campus &  3.84($\pm$1.67)     &  4.05($\pm$1.42)     &  4.76($\pm$2.06)     & N/A \bigstrut[t]\\
          & Grass land &   3.96($\pm$1.74)    &   2.41($\pm$1.32)  & 3.98($\pm$2.01)     & N/A \\
          & Basketball field &  6.74($\pm$3.15)     &  6.04($\pm$2.78)     &  8.66($\pm$3.18)     &  12.30($\pm$3.84) \bigstrut[b]\\
    \hline
    \multirow{3}[2]{*}{\texttt{Det+Our\_Ground\_Est} } & Campus &  2.22($\pm$1.12)     &  2.04($\pm$0.78)     &  2.61($\pm$1.47)    & N/A \bigstrut[t]\\
          & Grass land &  2.27($\pm$1.16)     &   1.15($\pm$0.77)    &  1.98($\pm$1.43)     & N/A \\
          & Basketball field &  3.21($\pm$1.84)     &  2.49($\pm$1.66)     &  4.47($\pm$2.12)    & 6.71($\pm$2.33) \bigstrut[b]\\
    \hline
    \multirow{3}[2]{*}{\textbf{\texttt{Det+Trk+Our\_Ground\_Est}} } & Campus & 0.49($\pm$0.31) & 0.47($\pm$0.08) & 1.21($\pm$0.54) & N/A \bigstrut[t]\\
          & Grass land &   0.78($\pm$0.31)    &   0.21($\pm$0.08)    &  0.94($\pm$0.35)     & N/A \\
          & Basketball field &   2.07($\pm$1.46)    &  1.97($\pm$1.22)     &  2.42($\pm$1.74)     &  3.87($\pm$1.95) \bigstrut[b]\\
    \hline
    \end{tabular}%
  \label{tab:ablation}%
\end{table*}%

\subsection{Ablation Study}
\label{sec:ablation_study}
\begin{figure}[t]
\begin{center}
\includegraphics[width=\linewidth]{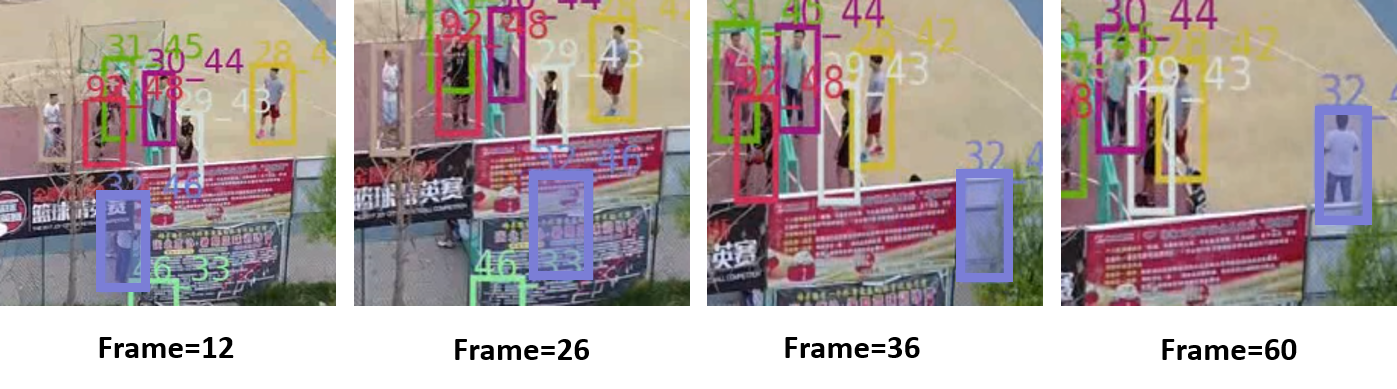}
\end{center}
   \caption{An example showing occlusion handling in testing sequence (basketball field). The trajectory of the person with a purple bounding box (ID: 32) recovers after fully occulsion. }
\label{fig:occu}
\end{figure}

\paragraph{\textbf{Occlusion Handling.}}
Better occlusion handling can help improving the 3D object localization performance. When an object is occluded, the detection is very likely to be unreliable or missing, which can generate wrong 2D bounding boxes or even no bounding boxes at all. The TNT tracker can handle the partial and full occlusions for a long duration. In Figure~\ref{fig:occu} of the basketball sequence, the person with a purple bounding box is fully occluded by a billboard from frame 12, but its trajectory is recovered after it appears again at frame 40.

\paragraph{\textbf{Ground Plane Estimation and Tracking.}}
To demonstrate the effectiveness of each of the our modules, we show the object localization performance with different methods in Table~\ref{tab:ablation}. \texttt{Det+Flat\_ Ground\_Asmp} denotes performing detection only and assuming a flat ground plane, i.e., unit normal vector of ${[0,-1,0]^T}$. \texttt{Det+Our\_ Ground\_Est} uses our ground plane estimation method in \ref{sec:depth_multi_view}. Note that the localization performance is especially improved for far objects, since small errors in ground plane can have a large impact on error over longer distances. Finally, in \texttt{Det+Trk+Our\_Ground\_Est}, the tracking method is added for comparison. In TNT, the unweighted moving average algorithm is applied to adjust the size of the bounding box when unreliable detection occurs. If the detection score is below threshold (0.2), the size of the bounding box is then determined by the past $k$ frames. Let $\{\mathbf{s}_{i,t}\}$, where $i\in\{1,2,3,4\}$ be four corner points of the target bounding box in the $t$-th frame and $\{\mathbf{x}_{i,t}\}$ be the detection outputs. The recursive formula of the unweighted moving average is
\begin{equation}
    {s_{i,t}} = {s_{i,t-1}} + \frac{{{x_{i,t}} - {x_{i,t - k}}}}{k}
\end{equation}
It is observed that the error decreases further, since the localization can now be estimated on more reliable detection bounding boxes with the help of tracking.

\begin{figure}[t]
\begin{center}
\includegraphics[width=\linewidth]{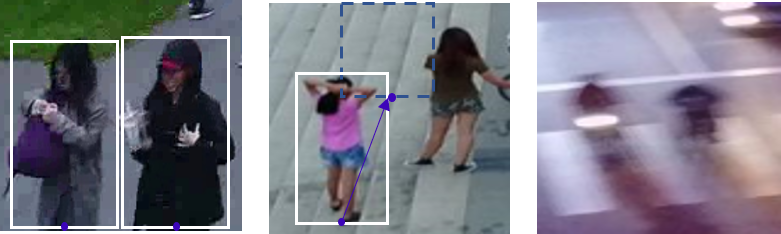}
\end{center}
  \caption{Typical issues (e.g. view of truncation, incorrect ground plane estimation and motion blur) that affect 3D localization performance.}
\label{fig:failure}
\end{figure}

\paragraph{\textbf{Failure Modes.}}
We illustrate some failure cases in Figure~\ref{fig:failure}, which includes field of view truncation that cause the bottom center of the bounding box no longer being the actual footpoint of the object. Failures can also occur due to incorrect ground plane estimation, and the abrupt camera motion with blurring.

\section{Conclusion and Future Study}
\label{sec:conclusion}
In this work, we have presented a novel framework for drone-based tracking and 3D object localization system. It combines CNN-based object detection, multi-object tracking, ground plane estimation and finally, 3D localization of the ground targets. Both the tracking performance and 3D localization performance are compared with either the state-of-the-art or ground truth. The robustness of our system is shown to handle most of the cases by drone, including occlusion handling and camera fast motions.

However, our work does have a few limitations. Although we demonstrate the fast camera motions may not affect the performance of tracking, it may affect the group plane estimation. When performing the epipolar search, it is not able to obtain the depth if the camera performs pure rotation, which is usually the case for the drone. A possible solution is to take the monocular depth map by CNN into considerations \cite{saxena2006learning}.  Since we are able to get 3D positions of each objects from proposed system, our future work also explores the 3D tracking so the trajectory will be much smoother compared to 2D. By adding some constraints into 3D trajectories, we believe the system will become more robust and effective.

\bibliographystyle{ACM-Reference-Format}
\balance
\bibliography{sample-base}

\end{document}